\documentclass[final]{cvpr}

\usepackage{times}
\usepackage{epsfig}
\usepackage{graphicx}
\usepackage{amsmath}
\usepackage{amssymb}
\usepackage{booktabs,tabularx}
\usepackage[labelfont=bf]{caption}
\usepackage{xparse}
\usepackage{multirow}
\usepackage{subfig}
\usepackage{bm}
\DeclareMathOperator{\Tr}{Tr} 
\DeclareMathOperator{\Log}{Log}

\NewDocumentCommand{\set}{o m}{%
  \IfNoValueTF{#1}
    {\{#2\}}
    {\{#1 \mid #2\}}%
}

\def\inv#1{{\color{blue}{\bf {w/ ours}}}}

\usepackage[pagebackref=true,breaklinks=true,colorlinks,bookmarks=false]{hyperref}
\usepackage{xr-hyper}
\makeatletter
\newcommand*{\addFileDependency}[1]{
  \typeout{(#1)}
  \@addtofilelist{#1}
  \IfFileExists{#1}{}{\typeout{No file #1.}}
}
\makeatother

\newcommand*{\myexternaldocument}[1]{%
    \externaldocument{#1}%
    \addFileDependency{#1.tex}%
    \addFileDependency{#1.aux}%
}
\myexternaldocument{supp}

\def\cvprPaperID{10853} 
\def\confYear{CVPR 2021}

\begin{document}

\title{InverseForm: A Loss Function for Structured Boundary-Aware Segmentation  }

\author{
Shubhankar Borse
\and
Ying Wang
\and
Yizhe Zhang
\and
Fatih Porikli
\and
Qualcomm AI Research\thanks{Qualcomm AI Research is an initiative of Qualcomm Technologies, Inc.}\\
{\tt\small \{sborse, yizhez, fporikli\}@qti.qualcomm.com, yingwang0022@gmail.com}\\
}

\maketitle

\begin{abstract}
We present a novel boundary-aware loss term for semantic segmentation using an inverse-transformation network, which efficiently learns the degree of parametric transformations between estimated and target boundaries. This plug-in loss term complements the cross-entropy loss in capturing boundary transformations and allows consistent and significant performance improvement on segmentation backbone models without increasing their size and computational complexity. We analyze the quantitative and qualitative effects of our loss function on three indoor and outdoor segmentation benchmarks, including Cityscapes, NYU-Depth-v2, and PASCAL, integrating it into the training phase of several backbone networks in both single-task and multi-task settings. Our extensive experiments show that the proposed method consistently outperforms baselines, and even sets the new state-of-the-art on two datasets.
\end{abstract}

\section{Introduction}

Semantic segmentation is a fundamental computer vision task with numerous real-world applications such as autonomous driving, medical image analysis, 3D reconstruction, AR/VR and visual surveillance. It aims to perform pixel-level labeling with a given set of target categories. 

There have been notable advancements in semantic segmentation thanks to recent solutions based on deep learning models, such as the end-to-end fully convolutional networks (FCN)~\cite{long2015fully} that lead to significant performance gains in popular benchmarks. Extensive studies have been conducted to improve semantic segmentation. One prevailing direction is to integrate multi-resolution and hierarchical feature maps~\cite{tao2020hierarchical}\cite{sun2019high}. Another ambitious objective is to exploit boundary information to enhance segmentation~\cite{yuan2020segfix}\cite{takikawa2019gated} further, driven by the observations that segmentation prediction errors are more likely to occur near the boundaries~\cite{takikawa2019gated}~\cite{pointrend}. In parallel, multi-task learning (MTL) frameworks~\cite{dvornik2017blitznet}\cite{maninis2019attentive} explored joint optimization of semantic segmentation and supplemental tasks such as boundary detection~\cite{dvornik2017blitznet}~\cite{vandenhende2020mti}\cite{maninis2019attentive} or depth estimation~\cite{xu2018pad}.

\begin{figure}[t]
\center
\begin{tabularx}{\textwidth}{c c c}
\hspace{-4mm}
\includegraphics[ width=0.33\linewidth, keepaspectratio]{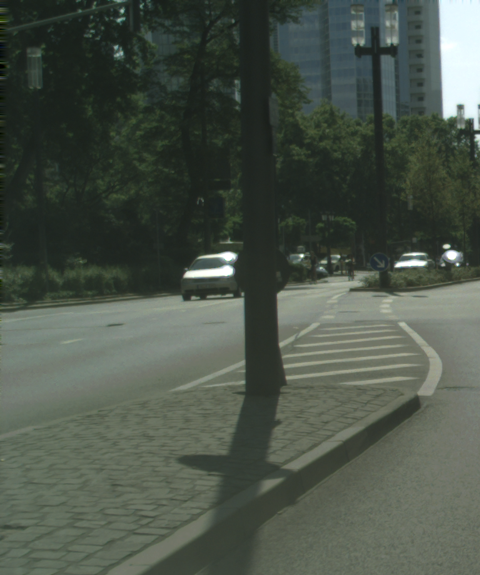} &
\hspace{-4mm}
\includegraphics[ width=0.33\linewidth, keepaspectratio]{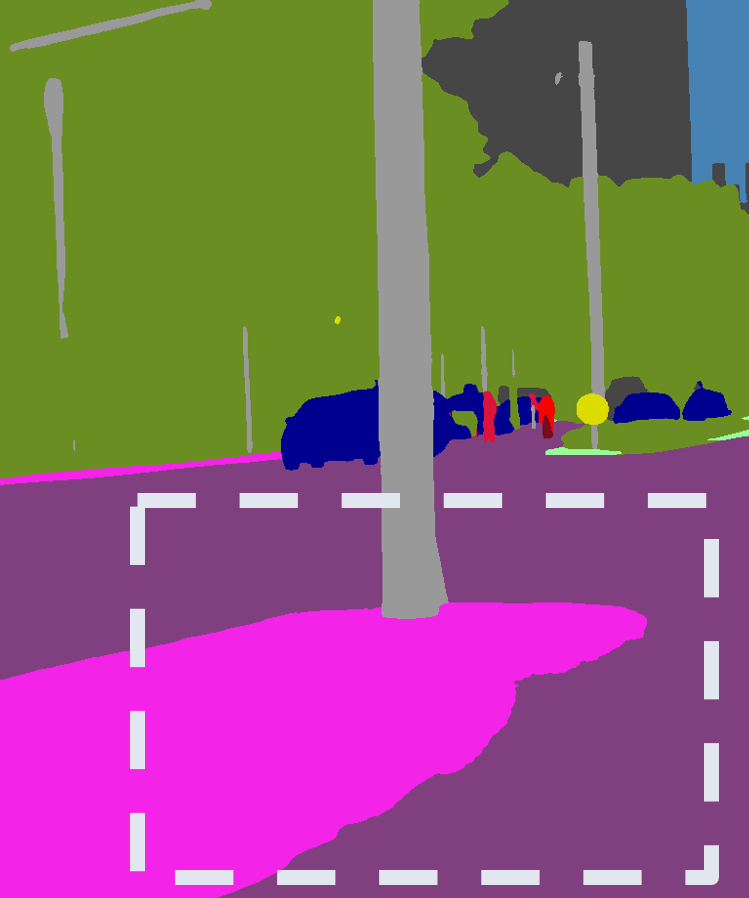} &
\hspace{-4mm}
\includegraphics[ width=0.33\linewidth, keepaspectratio]{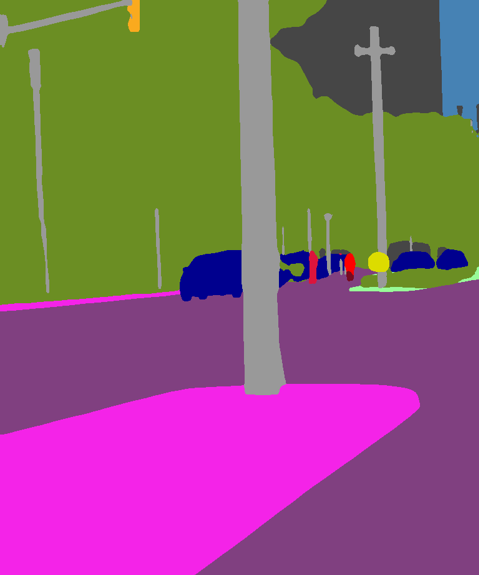}  \\
\hspace{-4mm}
\includegraphics[ width=0.33\linewidth, keepaspectratio]{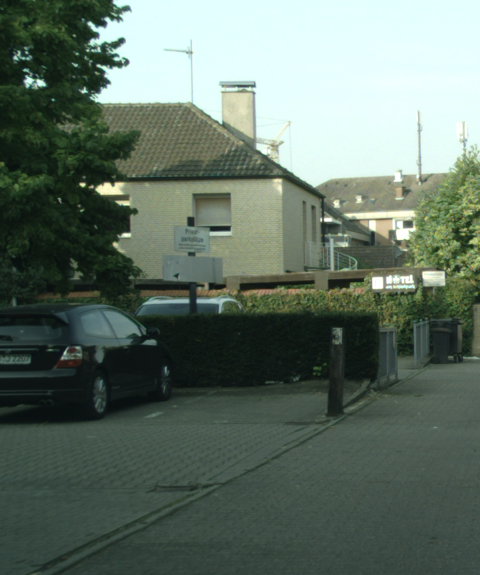}  &
\hspace{-4mm}
\includegraphics[ width=0.33\linewidth, keepaspectratio]{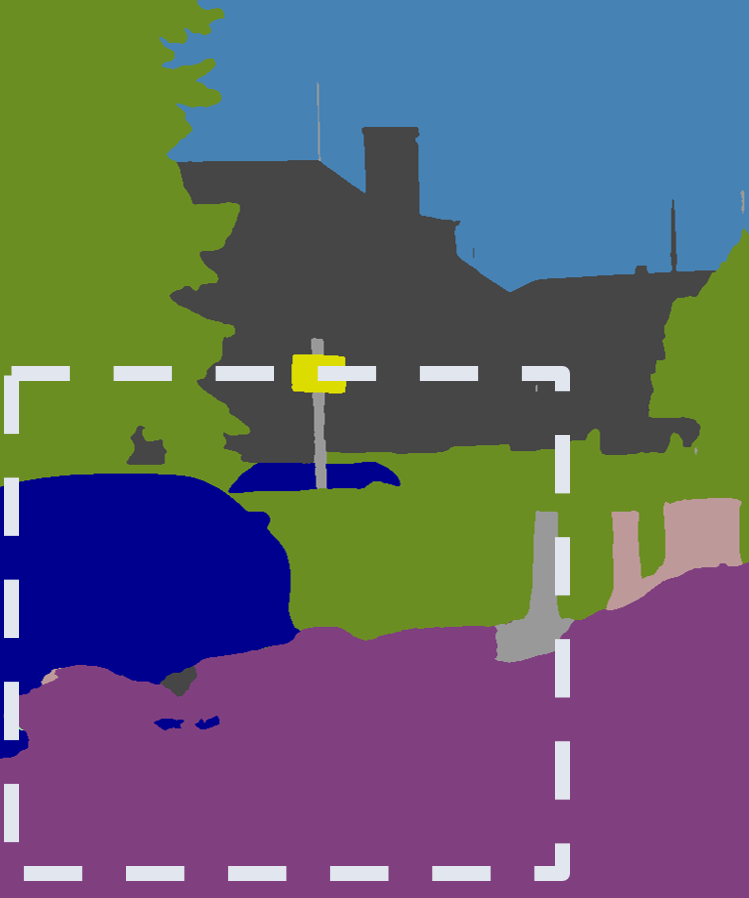}  &
\hspace{-4mm}
\includegraphics[ width=0.33\linewidth, keepaspectratio]{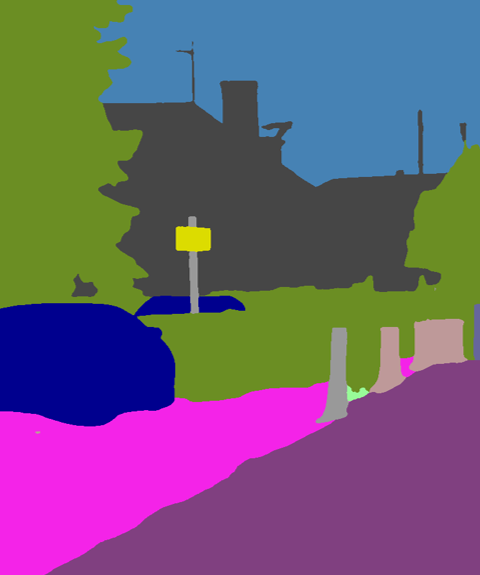}  \\
\hspace{-4mm}
\textbf{Image} & \hspace{-3mm} \textbf{Baseline} & \hspace{-3mm} \textbf{Baseline \inv{}} \\
\hspace{-2mm}

\end{tabularx}
\vspace{-2mm}
\caption{ Left: Images from Cityscapes \textit{val} benchmark~\cite{cordts2016cityscapes}. Middle: Segmented prediction for an HRNet-48-OCR~\cite{wang2020deep} ~\cite{yuan2020objectcontextual} baseline. Right: Same backbone trained using our InverseForm boundary loss. Our model achieves clear improvements, e.g. the curbside boundary in the top figure is aligned better with the structure of boundary, and the curbside in the bottom is correctly detected. 
} 
\label{fig:firstpage_cityscapes}
\end{figure}

Our approach is aligned with making the best use of boundary exploration. One of the main differences is that most of the previous works use a weighted cross-entropy loss as their loss function for boundary detection, which we show in Figure~\ref{fig: affine_transform}, is sub-optimal for measuring the boundary shifts. This is also partially observed in~\cite{kokkinos2016pushing}. The cross-entropy loss mainly builds on estimated and ground-truth pixel label changes but ignores the spatial distance of the pixels from the target boundaries. It cannot effectively measure localized spatial variations such as translation, rotation, or scaling between predicted and target boundaries. 

Motivated to address this limitation, we introduce a boundary distance-based measure called InverseForm, into the popular segmentation loss functions. We design an inverse transformation network to model the distance between boundary maps, which can efficiently learn the degree of parametric transformations between local spatial regions. This measure allows us to achieve a significant and consistent improvement in semantic segmentation using any backbone model without increasing the inference size and computational complexity of the network. 

Specifically, we propose a boundary-aware segmentation scheme during the training phase, by integrating our spatial-distance loss, InverseForm, into the existing pixel-based loss. Our distance-based loss complements the pixel-based loss in capturing boundary transformations. We utilize our inverse transformation network for distance measurement from boundaries and jointly optimize pixel-label accuracy and boundary distance. We can integrate our proposed scheme into any segmentation model; for instance, we adopt the latest HRNet~\cite{wang2020deep} architecture as one of the backbones since it maintains high resolution feature maps.

We also adopt various MTL frameworks~\cite{maninis2019attentive}~\cite{vandenhende2020mti} to leverage on their boundary detection task towards further segmentation performance improvement, at no added computational and memory cost. In this variant, we show a consistent performance improvement across all tasks as a plus.

We conduct comprehensive experiments on large benchmark datasets including NYU-Depth-v2~\cite{silberman2012indoor}, Cityscapes~\cite{cordts2016cityscapes} and PASCAL-Context~\cite{pascal2014}. For NYU-Depth-v2, we show that InverseForm based segmentation outperforms the state-of-art in terms of mean intersection-over-union (mIoU). We also show that we outperform state-of-the-art Multi-task learning methods on PASCAL in terms of their multi-task performance. This includes superior performance in predicted edge quality on odsF-score~\cite{odsf} and improvement in other tasks, such as human-parts estimation and saliency estimation, on mIoU. Then, we perform rigorous experiments on Cityscapes to compare our method with contemporary works such as SegFix~\cite{yuan2020segfix} and Gated-SCNN~\cite{takikawa2019gated}. 

The contributions of our work include the following:
\begin{itemize}

\item We propose a boundary distance-based measure, InverseForm, to improve semantic segmentation. We show that our specifically tailored measure is significantly more capable of capturing the spatial boundary transforms than cross-entropy based measures, thus resulting in more accurate segmentation results. 

\item Our scheme is agnostic to the backbone architecture choice and is very flexible to be plugged into any existing segmentation model, with no additional inference cost. It does not impact the main structure of the network due to its plug-and-play property. It is flexible and can fit into multi-task learning frameworks. 

\item We show through extensive experiments that our boundary-aware-segmentation method consistently outperforms its baselines, and also outperforms the state-of-the-art methods in both single-task (on NYU-Depth-v2) and multi-task settings (on PASCAL).

\end{itemize}

\section{Related Work}

\textbf{Semantic segmentation:} The introduction of fully convolutional networks (FCNs)~\cite{long2015fully} led to significant progress in semantic segmentation. Many works build on FCNs to improve segmentation performance. The work in \cite{krahenbuhl2011efficient} proposed an approximate inference algorithm for conditional random fields (CRFs) models to improve segmentation efficiency. Various architectures have been proposed to improve inference speed, such as DeepLab~\cite{chen2017deeplab}, PSPNet~\cite{zhao2017pyramid}, and HRNet~\cite{wang2020deep}. Numerous recent works utilize HRNet by integrating diverse contextual models~\cite{yuan2020objectcontextual}~\cite{yuan2020segfix}~\cite{takikawa2019gated}.

\textbf{Boundary-aware segmentation:} Boundary-aware segmentation has been studied in quite a few recent efforts. There are several ways of modeling this problem. In~\cite{bertasius2016semantic}, a global energy model called boundary neural field (BNF) is introduced to predict boundary cues to enhance semantic segmentation. The authors show that the energy decomposes semantic segmentation into a set of binary problems which can be relaxed and solved by global optimization. In~\cite{ding2019boundary} a boundary-aware feature propagation module (BFP) is proposed to propagate local features isolated by the boundaries learned in the UAG-structured image. The work in~\cite{liu2017learning} learns semantically-aware affinity through a spatial propagation network to refine image segmentation boundaries. Similarly, \cite{chen2016semantic} adopts a fast domain transform filtering on the output and generates edge maps to refine segmentation output. Gated-SCNN \cite{takikawa2019gated} injects the learned boundary into intermediate layers to improve segmentation. This is done using gating to allow interaction between the main segmentation stream and the shape stream. However, these boundaries are learnt using a pixel-based loss. We show in Section~\ref{sec:experiment} how this loss benefits from using our proposed spatial loss. Other related works include~\cite{pointrend}~\cite{ding2019semantic}~\cite{ding2018context}~\cite{lin2017refinenet}~\cite{cheng2017fusionnet} \cite{peng2017large}, and~\cite{kokkinos2016pushing}. 

One of the main differences between our work and all the previous works is that most of the previous works use weighted cross-entropy as their loss function for boundary detection, which is sub-optimal for measuring the boundary changes. The cross-entropy loss mainly considers the pixel label but ignores the distance between the boundaries. It cannot effectively measure the shifts, scales, rotations, and other spatial changes between two image regions. In our work, we introduce a distance-based metric into this loss function. Specifically, we employ an inverse-transformation network to model the spatial distance between boundary maps, which is shown to significantly improve the capability in capturing such boundary transforms. An interesting recent work which makes use of spatial relations is SegFix~\cite{yuan2020segfix}. It learns discrete distance-transformation maps to incorporate spatial relations. Instead of learning these offset maps, our work regresses over homography parameters over which it uses a derived distance computation. We show both qualitatively and quantitatively that this is more effective in Section~\ref{sec:experiment}. Another major difference between our work and all previous mentioned works is that our proposed method requires no extra cost during inference.

\textbf{Multi-task learning:} Multi-task learning (MTL)~\cite{kokkinos2016ubernet} learns shared representations from multi-task supervisory signals under a unified framework. It is effective at exploring interactions among multiple tasks while saving memory and computation. \cite{vandenhende2020mti} shows the superior performance of multi-task learning for semantic segmentation and depth estimation for NYU-Depth-v2. \cite{vandenhende2020multitask} provides a comprehensive overview of multi-task learning techniques. In~\cite{maninis2019attentive} the authors address the task interference issue by performing one task at a time using the multi-task framework. They allow the network to adapt its behavior through task attention.

\section{Proposed Scheme: InverseForm}
Below, we explain our method for boundary-aware segmentation using a spatial distance-based loss function.

\begin{figure}[t]
    \centering
     \includegraphics[ width=0.44\textwidth, keepaspectratio]{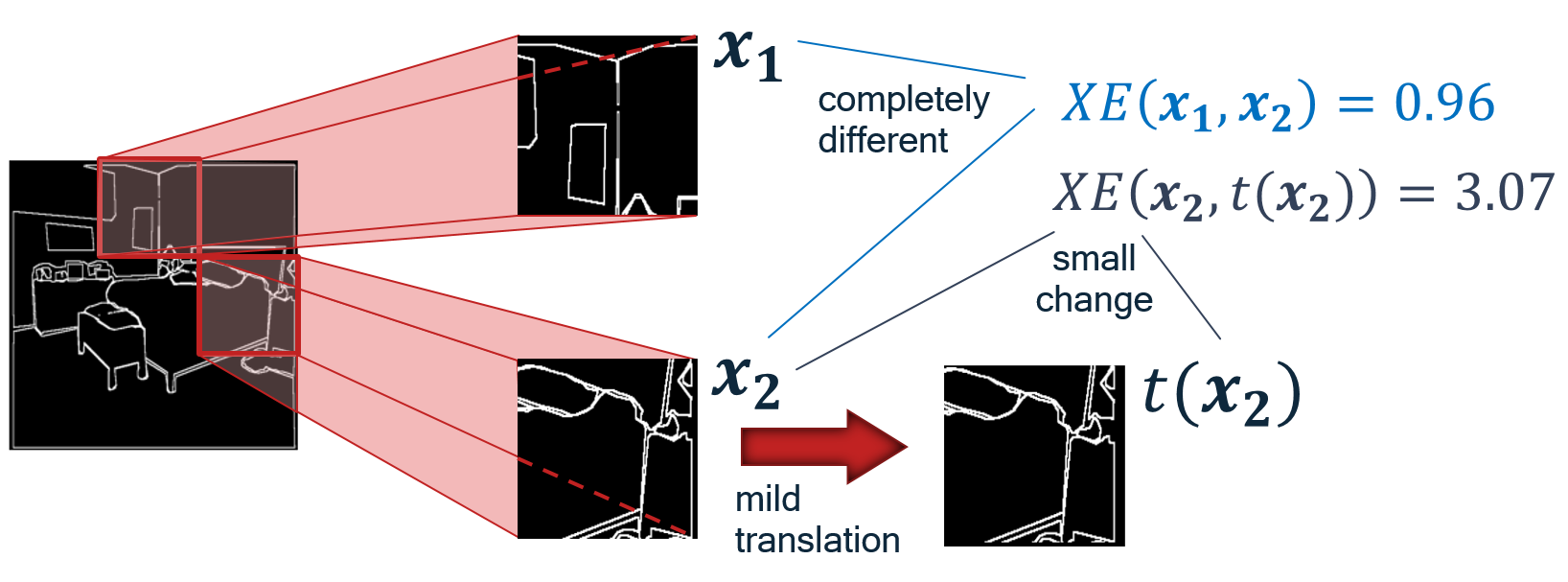}
 \captionof{figure}{ Cross-entropy(XE) based distance fails for spatial transformations of boundaries. In the above example, \(x_{1}\) and \(x_{2}\) are slices sampled from a boundary map. Only a mild shift is applied to \(x_{2}\) to generate a third slice \(t(x_{2})\). 
 } \label{fig: affine_transform}
\end{figure}

\subsection{Motivation for distance-based metrics}

Boundary-aware segmentation approaches use pixel-wise cross-entropy or balanced cross-entropy losses. Such loss functions take into account the pixel-wise features (intensity, etc.) but not spatial distance between object boundaries and ground-truth boundaries. Hence, they are insufficient for imposing boundary alignment for segmentation. 

This drawback is illustrated in Figure \ref{fig: affine_transform}. We pick an image from NYU-Depth-v2 and convert the ground-truth segmentation into a boundary map. We then sample two regions \(\boldsymbol{x_{1}}\) and \(\boldsymbol{x_{2}}\) from the map. We apply a spatial translation by 3 pixels in both dimensions to \(\boldsymbol{x_{2}}\). The cross-entropy(XE) between pairs of these images is also shown. Notice that the actual difference between the region \(\boldsymbol{x_{2}}\) and its transformed version \(t(\boldsymbol{x_{2}})\) is not high, as the boundary maps are shifted by only a few pixels. However, cross-entropy between this pair of images is much higher than the cross-entropy between \(\boldsymbol{x_{1}}\) and \(\boldsymbol{x_{2}}\), which have no correlation. Ideally, if \(\boldsymbol{x_{2}}\) is the ground truth map; a network should be penalized more when it generates \(\boldsymbol{x_{1}}\) as compared to when it generates \(t(\boldsymbol{x_{2}})\). This is clearly not the case. Pixel-based losses are thus, not enough to measure `distance' between such inputs. To counter the effect of sensitivity to small displacements, boundary detection networks trained with pixel-based losses produce thicker and distorted boundaries. Hence, there is an existing need to measure a distance function between two boundary maps which models possible spatial transformations between them.

Some works use Hausdorff distance~\cite{hausdorff1} to model this measure between boundaries for instance-level segmentation, but this loss cannot be efficiently applied in a semantic segmentation setting. The correlation operation could also be considered between two images, but this would only model translations between image boundaries and cannot model other transformations. Thus, we need a loss function that accurately models the distance between two images of object boundaries and can be computed efficiently.

\subsection{Inverse transformation network}\label{sec:inv_transf_network}

\begin{figure}[t]
\centering
\subfloat[Spatial transformer network]{
  \includegraphics[ width=0.9\linewidth, keepaspectratio]{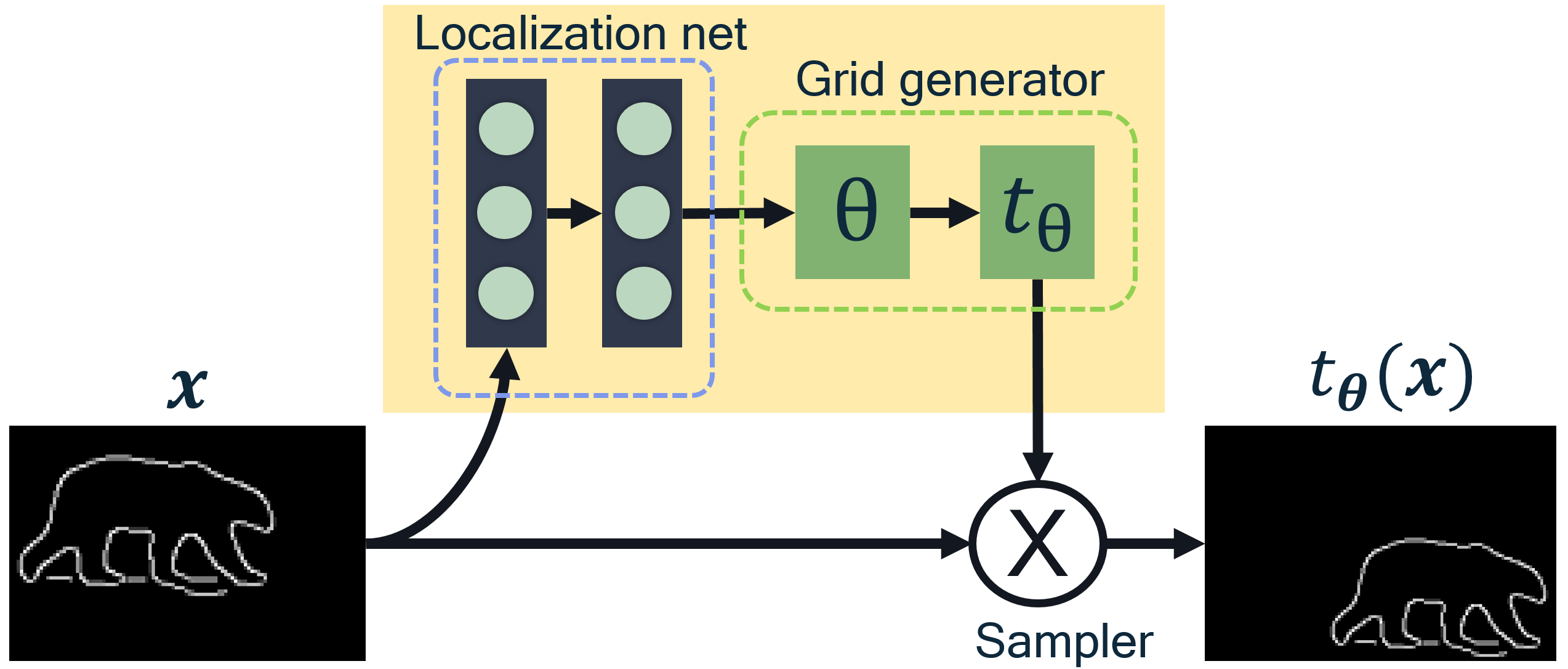}}
  \vspace{0.2cm}
\subfloat[Inverse transformation network]{
  \includegraphics[ width=0.9\linewidth, keepaspectratio]{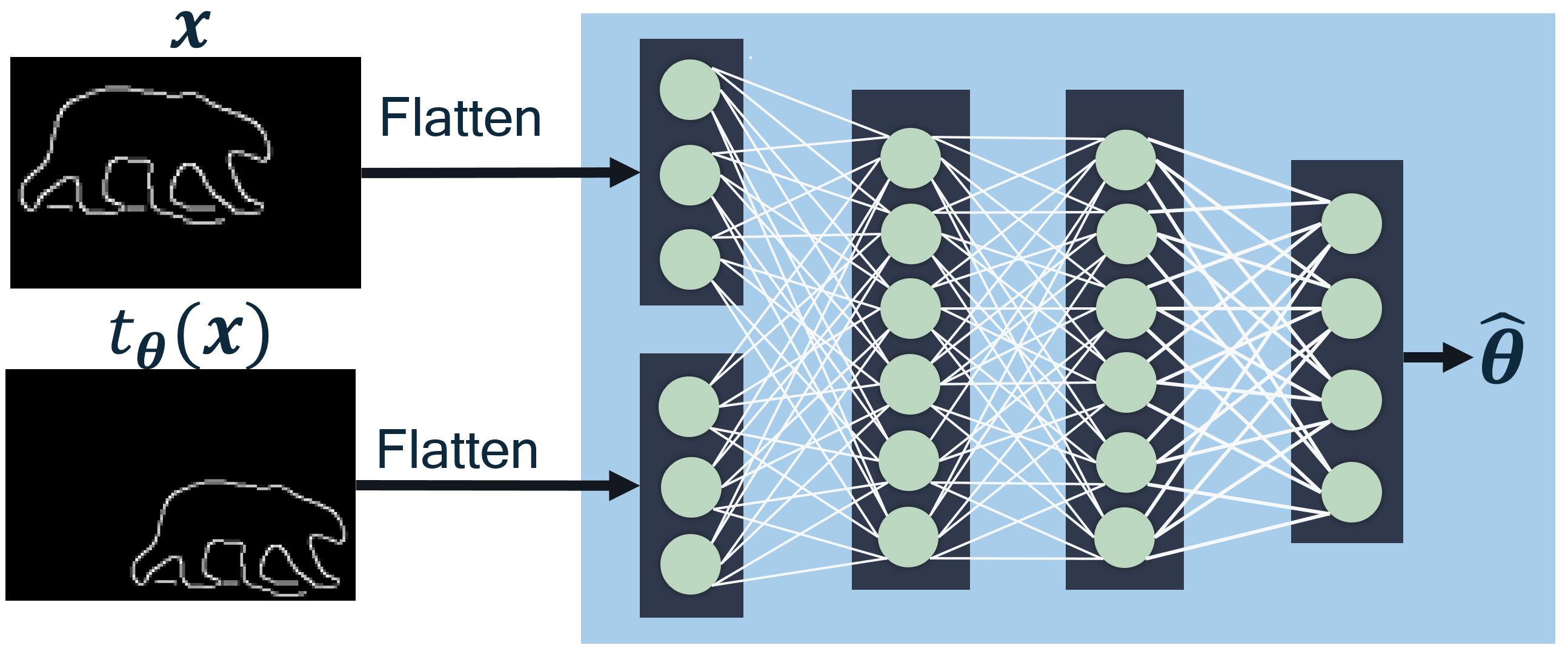}}
  \caption{Spatial transformer (a) and our inverse-transformation network (b). Our inverse-transformation network architecture inputs two images \(\boldsymbol{x}\) and \(t_{\boldsymbol{\theta}}(\boldsymbol{x})\), and predicts the transformation parameters \(\hat{\boldsymbol{\theta}}\). This performs the inverse operation as that of the spatial transformer network. 
  }
  \label{fig:affine_transform}
\end{figure}

To model spatial relations between two boundary maps, we assume that they are related to each other through a homography transformation. Thus, we create a network that inputs two boundary maps and predicts the ``homography change'' as its output. Theoretically, this network performs the inverse operation of a spatial transformer network (STN)~\cite{spatialtx}. As shown in Figure~\ref{fig:affine_transform}(a), STN takes an image \(\boldsymbol{x}\) as its input, generates a controlled transformation matrix $\boldsymbol{\theta}$, and produces a realistic transformed boundary image \(t_{\boldsymbol{\theta}}(\boldsymbol{x})\). 

Figure~\ref{fig:affine_transform}(b) shows our proposed network. Inputs to our network are two boundary maps. The network regresses on the transformation parameters between these maps as its output. Thus, we call it an \emph{inverse transformation network}. Some recent works~\cite{aetv2}~\cite{zhang2019aet} attempted to train a similar network to use its encoder for an unsupervised learning task. Our architecture and application are completely different; we use the network to model a spatial distance measure, and our model has only dense layers. We compare our inverse transformation network's performance with~\cite{aetv2} in Section~\ref{sec:ablation}, where we discuss the translation-invariance of convolutions and why we selected a dense architecture.

The outputs of the inverse-transformation network are the coefficients of the homography matrix. There are numerous methods to formulate a distance metric from these values. Below, we present two distance metric choices. 

We note that one may also attempt to directly regress on the distance instead of estimating the transformation coefficients. However, such an approach would not allow optimization of the boundary-aware segmentation network.

\subsection{Measuring distances from homography}\label{sec:distancemeasures}

Assuming two boundary maps are related to each other with a homography transformation, our inverse-transformation net inputs two boundary maps and estimates the transformation matrix parameters. Once our inverse-transformation network is trained, we freeze its weights to generate our proposed InverseForm loss. If there is a perfect match between input boundary maps, the network should estimate an identity matrix. Thus, a measure of spatial distance between the boundary maps can be calculated by comparing the network outputs to an identity matrix. 

Here are two distance measures we propose to relate the two matrices: Euclidean distance and Geodesic distance. 

\textbf{Euclidean distance:} The Euclidean distance between two homography matrices is a measure of spatial distance between the two input boundary maps. We observed that Euclidean distance could model shift and scale relations well. However, it fails to reflect rotations and other perspective transformations. Considering the output of the network as $\hat{\boldsymbol{\theta}}$, the $3\times 3$ identity matrix is $\boldsymbol{I_{3}}$ and $\|\cdot \|_F$ is the Frobenius norm, the InverseForm distance is calculated as
\begin{equation}\label{equ:d_if}
   d_{if}(\boldsymbol{x}, t_{\boldsymbol{\theta}}(\boldsymbol{x})) = \left\Vert \hat{\boldsymbol{\theta}} - \boldsymbol{I_{3}} \right\Vert_{F}. 
\end{equation}

We train the inverse-transformation network by reducing the Euclidean distance between predicted and ground-truth parameters, and use Equation~(\ref{equ:d_if}) at inference time.

\textbf{Geodesic distance:} Homography transformations reside on an analytical manifold instead of a flat Euclidean space. The geodesic distance can capture these transformations, as explained in the recent works~\cite{aetv2}~\cite{liegroups}. Considering the ground-truth parameters as $\boldsymbol{\theta}$, we use the formulation of geodesic distance from both these works, as follows 
\begin{equation}\label{equ:d_if_geo}
   d_{if}(\boldsymbol{x}, t_{\boldsymbol{\theta}}(\boldsymbol{x})) = \left\Vert \frac{\Log(\boldsymbol{\theta}^{-1}\hat{\boldsymbol{\theta}})}{\Log(\boldsymbol{I_{3}})} \right\Vert_{F} .
\end{equation}
To train the network using this formulation of geodesic distance, we are required to calculate gradient errors over the Riemannian logarithm, which does not have a closed-form solution. Thus, we use the method in~\cite{aetv2} to project the homography Lie group onto a subgroup \textbf{SO(3)}~\cite{taylor1994minimization} where the calculation of geodesic distance does not need the Riemannian logarithm. The formulation is given by
\begin{equation}\label{equ:d_if_real_geo}
   d_{if}(\boldsymbol{x}, t_{\boldsymbol{\theta}}(\boldsymbol{x})) = \arccos\left[\frac{\Tr{(\boldsymbol{P})}-1}{2} \right] + \lambda \Tr{(\boldsymbol{R_{\pi}}^{T} \boldsymbol{R_{\pi}})}, 
\end{equation}
where $\lambda$ is a weighting parameter set at 0.1 and the projection $\boldsymbol{P}$ onto the rotation group \textbf{SO(3)} is given as 
\( 
   \boldsymbol{P} = \boldsymbol{U} diag\{1, 1, det(\boldsymbol{U}\boldsymbol{V})^{T}\}\boldsymbol{V}^{T} 
\) 
and the projection residual $\boldsymbol{R_{\pi}}$ is calculated as
\begin{equation}\label{equ:r_if_real_geo}
   \boldsymbol{R_{\pi}} = \boldsymbol{\theta}^{-1}\hat{\boldsymbol{\theta}} - \boldsymbol{P}.
\end{equation}
This formulation of geodesic distance can be used to train the inverse-transformation network. During inference, we insert $\boldsymbol{\theta} = \boldsymbol{I_{3}}$ in Equation~(\ref{equ:r_if_real_geo}) to compute the distance between the two inputs to the inverse-transformation network.

For using geodesic distance, there are 8 degrees of freedom in the $3\times3$ homography matrix. The inverse-transformation network predicts 8 values and we set the 9th value to 1. 
On the other hand, the matrix has 6 degrees of freedom if we assume only 2D affine transformations. Hence, the network must predict 6 values if we use the Euclidean distance measure. We have compared the effect of using these distance measures in Section~\ref{sec:ablation_architecture_search}.

\begin{figure*}[t]
\centering
 \includegraphics[ width=0.99\textwidth,   keepaspectratio]{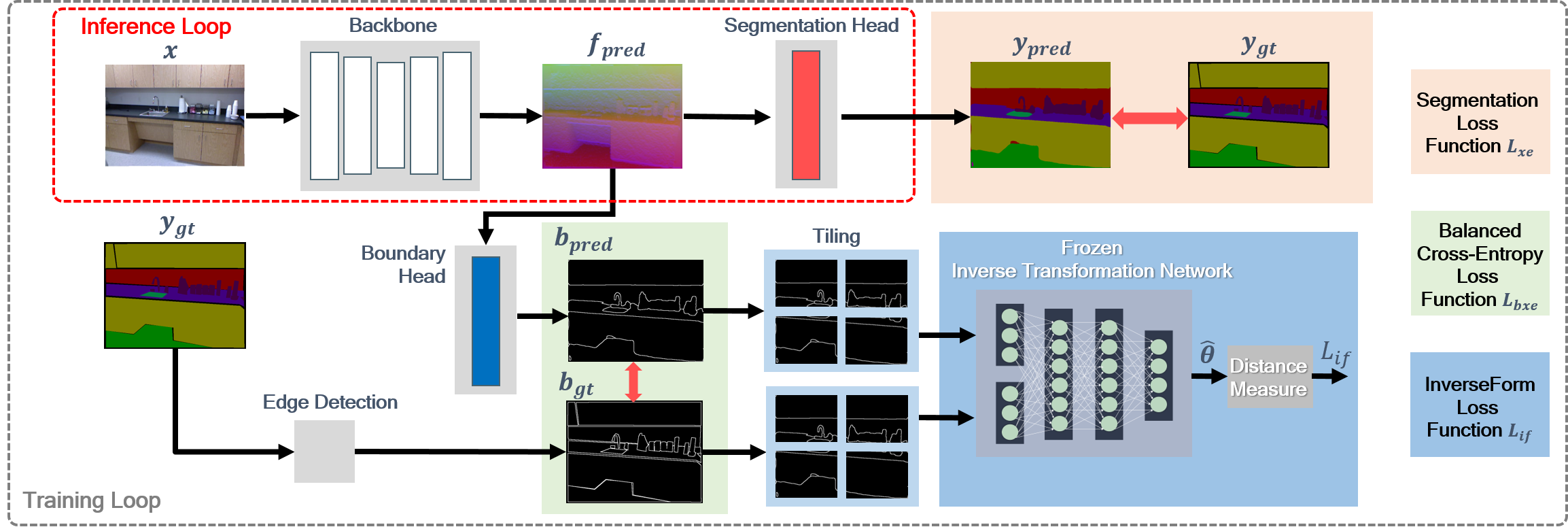}
\caption{Overall framework for our proposed boundary-aware segmentation. } \label{fig:overall_framework}
\end{figure*}

\subsection{Using InverseForm as a loss function}

We first train the inverse-transformation network using boundary maps of images sampled from the target dataset. We apply the STN~\cite{spatialtx} to generate the transformed versions of boundary images. We could also do this by randomly sampling transformation parameters using the method described in~\cite{aetv2}, however this leads to lesser realistic transformations. Then, we use these images and their affine transformations as inputs to the inverse-transformation network. Before feeding boundary maps to the network, we split the images into smaller tiles. Tiling is done in a way that the local context can be preserved along with the global context. Ideally, the best tiling dimension should provide a balance between local and global contexts. We discuss the effect of tiling dimension in the Appendix. Training details and hyperparameters used can be found in the Appendix.

Once we train the inverse-transformation network, we freeze its weights and use it as a loss function in a boundary-aware segmentation setting. Using networks as loss functions is a common practice in generative adversarial network literature ~\cite{goodfellow2014generative}. However, the distance computed based on inverse-transformation network outputs doesn't fall in the category of discriminator or adversarial losses, according to the definition given by ~\cite{adversarialdef}.

To model the loss function, we assume the predicted boundary \(\boldsymbol{b_{pred}}\) is a transformed version of the ground truth boundary label \(\boldsymbol{b_{gt}}\). i.e. \(\boldsymbol{b_{pred}} = t_{\boldsymbol{\theta}}(\boldsymbol{b_{gt}}).\)
Hence, we can formulate our InverseForm loss function in terms of the spatial distance calculated in Section~\ref{sec:bd_aware_seg}. This function first splits the input boundaries \(\boldsymbol{b_{pred}}\) and \(\boldsymbol{b_{gt}}\) into $N$ smaller tiles \(\boldsymbol{b_{pred, j}}\) and \(\boldsymbol{b_{gt, j}}\). Next, it passes the inputs through the inverse-transformer network and calculates spatial distance in one of the ways described in Section~\ref{sec:inv_transf_network}. The formulation of our InverseForm loss function \(L_{if}\) is given by
\begin{equation}\label{eq:L_if}
L_{if}(\boldsymbol{b_{pred}}, \boldsymbol{b_{gt}}) = \sum_{j=1}^{N} d_{if}(\boldsymbol{b_{pred, j}}, \boldsymbol{b_{gt, j}}) \;.
\end{equation}

\subsection{Boundary-aware segmentation setup}\label{sec:bd_aware_seg}

We train various architectures using our loss function, with both single-task and multi-task settings in Section~\ref{sec:experiment}. For multi-task settings, we directly use the architecture used by MTI-Nets~\cite{vandenhende2020mti} and ASTMT~\cite{maninis2019attentive} and add our InverseForm loss term to the boundary loss.
 
To train single-task architectures using InverseForm loss, we use a simple boundary-aware segmentation setup, as displayed in Figure~\ref{fig:overall_framework}. This setup could be used over any backbone. Consider \(\boldsymbol{x}\) is the input image, \(\boldsymbol{y_{gt}}\) is the ground-truth segmentation map and \(\boldsymbol{y_{pred}}\) is the network prediction. The network produces intermediate features \(\boldsymbol{f_{pred}}\), which are fed into a segmentation backbone to produce the output segmentation. We pass the features \(\boldsymbol{f_{pred}}\) into an auxiliary head to produce a boundary map \(\boldsymbol{b_{pred}}\). Optionally, this boundary map could be concatenated with the features \(\boldsymbol{f_{pred}}\) and then fed to the segmentation head to produce the output \(\boldsymbol{y_{pred}}\). We don't add this supervision in our experiments to ensure zero computational overhead.

Pixel-wise cross-entropy loss \(L_{xe}\) is used on the segmentation output. The ground-truth \(\boldsymbol{y_{gt}}\) is passed through a Sobel filter~\cite{canny1986computational} to produce a binary boundary map. We integrate the InverseForm loss and weighted cross-entropy loss \(L_{bxe}\) over the predicted boundary and ground-truth maps into the overall loss function. Specifically, we define the overall loss function as  
\begin{align}\label{equ:L_total}
L_{total} = & \; L_{xe}(\boldsymbol{y_{pred}}, \boldsymbol{y_{gt}}) + \beta L_{bxe}(\boldsymbol{b_{pred}}, \boldsymbol{b_{gt}}) \nonumber\\
&+ \gamma L_{if}(\boldsymbol{b_{pred}}, \boldsymbol{b_{gt}})
\end{align}
where $L_{xe}$ is the cross-entropy loss, $L_{bxe}$ is weighted cross-entropy loss, and $L_{if}$ is the InverseForm loss defined in~\eqref{eq:L_if}. \(\beta, \gamma\) are the scaling parameters for the weighted cross-entropy loss and InverseForm loss terms, respectively. They are treated as hyperparameters and we discuss the effects of these terms in the Appendix.

\section{Experimental Results}\label{sec:experiment}

\noindent\textbf{Datasets:} We present performance scores for various tasks on three datasets, NYU-Depth-v2~\cite{silberman2012indoor}, PASCAL~\cite{pascal2014} and Cityscapes~\cite{cordts2016cityscapes}. We use the original 795 training and 654 testing images for NYU-Depth-v2. For PASCAL, we use a split from PASCAL-Context, which has annotations for segmentation, edge-detection, and human parts. We also use labels for surface normals and saliency from~\cite{maninis2019attentive}. This split for PASCAL is used by ASTMT~\cite{maninis2019attentive} and MTI-Nets~\cite{vandenhende2020mti} for a multi-task framework. For Cityscapes, we use their 2975/500/1525 train/val/test splits to report performance. Models reporting on test set are trained using train+val set.

\noindent\textbf{Evaluation metrics:} For single-task learning settings, we measure semantic segmentation performance. This is evaluated using mean intersection-over-union (mIoU) and pixel-accuracy. To show improvement on boundary regions, we use our implementation of multi-class Mean Boundary Accuracy(mBA) from~\cite{cheng2020cascadepsp}. This is computed over segmentation outputs. 
For multi-task learning settings on PASCAL, the model is trained for five tasks: semantic segmentation, boundary detection, saliency estimation, human parts and surface normal estimation. We measure mIoU for semantic segmentation, human parts, and saliency estimation. The mean error in predicted angles is calculated for surface normal, and the optimal dataset F-measure is used to evaluate boundary detection. We measure depth and semantic segmentation for NYU-Depth-v2. Depth estimation is measured using root mean square error (rmse). Finally, the multi-task performance of model $m$, \(\Delta_{m}\)~\cite{maninis2019attentive} is given by the relative increase wrt. single-task performance for all tasks:
\( \Delta_{m} = \frac{1}{T}\sum_{i=1}^{T} (-1)^{l_{i}}\frac{(M_{m,i}-M_{b,i})}{M_{b,i}} \).
Here, \(M_{m,i}\) and \(M_{b,i}\) are metrics for the multi-task model and single-task model for task $i$. $l_{i}$ is 1 when the metric is positive v/s 0 when the metric is negative.

\begin{table}[t] 
\centering
 \begin{tabular}{c|c|cc} 
 \hline
 Network & Tasks &  mIoU & mBA \\ 
 \hline
 \hline
  HRNet-w18 & S & 33.18 & 37.46 \\ 
  HRNet-w18 \inv{} & S+E & \textbf{34.79} & \textbf{40.72} \\ 
 \hline
  PAD-HRNet18 & S+D & 32.80 & 38.10 \\   
  PAD-HRNet18 & S+D+E+N & 33.10 & 42.69 \\   
  PAD-HRNet18 \inv{} & S+D+E+N & \textbf{34.70} & \textbf{43.24} \\   
 \hline
  MTI-HRNet18 & S+D & 35.12 & 42.44 \\
  MTI-HRNet18 & S+D+E+N & 37.49 & 43.26 \\
  MTI-HRNet18 \inv{} & S+D+E+N & \textbf{38.71} & \textbf{45.59} \\
 \hline
  HRNet-w48 & S & 45.70 & 56.01 \\ 
  HRNet-w48 \inv{} & S+E & \textbf{47.42} & \textbf{59.34} \\
 \hline
\end{tabular}
\caption{Comparing baselines for NYU-Depth-v2 using HRNet-w18 and HRNet-w48 backbones in both single task and multi-task learning using our loss. In multi-task settings, S:Semantic segmentation, D:Depth, E:Edge detection, N:Surface normal estimation. Consistent improvement in segmentation mIoU and boundary mBA is visible.}\label{tab:nyud}
\end{table}

\subsection{Results on NYU-Depth-v2}
\begin{figure}[t]
\begin{tabularx}{\textwidth}{cc}
    \hspace{-4mm}
     \includegraphics[ width=0.5\linewidth, keepaspectratio]{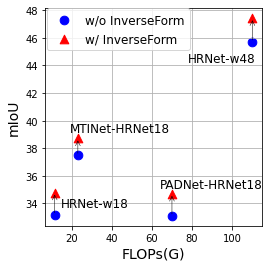}

     \includegraphics[ width=0.5\linewidth, keepaspectratio]{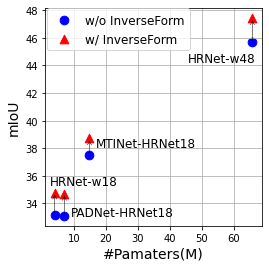}
\end{tabularx}

\caption{Comparison of mIoU v/s FLOPs and mIou v/s \# params for different schemes on NYU-Depth-v2 dataset.}\label{fig:miou_flops_nyud}
\end{figure}

For NYU-Depth-v2, we report scores using three backbone architectures, i.e. HRNet-w18, HRNet-w48 (adapted from~\cite{hrnet}) and ResNet-101~\cite{resnet}. For the vanilla HRNet-w18 and HRNet-w48 backbones, we train with InverseForm loss using the method described in Figure~\ref{fig:overall_framework}. We also add the InverseForm loss term to the boundary loss function within available multi-task baselines as implemented by MTI-Net~\cite{vandenhende2020mti} over HRNet-18. We used their implementation of PAD-Net~\cite{xu2018pad} in multi-task settings. These results are given in Table~\ref{tab:nyud}. For computational analysis, we plot the no. of parameters v/s mIoU and GFLOPs v/s mIoU curves in Figure~\ref{fig:miou_flops_nyud}. We remove the boundary detection head before running inference. Hence, our method increases the mIoU score without any extra computations during inference.  

The best scores reported on NYU-Depth-v2 are achieved by the multi-scale inference on RGBD inputs using SA-Gates~\cite{chen2020bidirectional} over ResNet-101 backbone and DeepLab-V3+~\cite{chen2017deeplab} as the decoder. We use their implementation and train this model with our InverseForm loss and the boundary-aware framework shown in Figure~\ref{fig:overall_framework}. Our model outperforms the SA-Gates baseline by 0.7\% mIoU and other state-of-the-art models by a large margin on NYU-Depth-v2, as shown in Table~\ref{tab:nyud_1}. We provide the hyperparameters for our experiments and the multi-task results for this experiment in the Appendix.

\begin{table}[t]

  \begin{tabular}{cc| cc} 
 \hline
 Network & Backbone & mIoU & Pixel-acc \\ 
 \hline
 \hline
  TD-Net~\cite{hu2020temporally} & PSPNet-50 & 43.5 & 55.2 \\ 
  PAD-Net~\cite{xu2018pad} & ResNet-50 & 50.2 & 75.2 \\   
  PAP-Net~\cite{zhang2019patternaffinitive} & ResNet-50 & 50.4 & 76.2 \\  
  MTI-Net~\cite{vandenhende2020mti} & HRNet-w48 & 49.0 & - \\   
  RSP~\cite{kong2017recurrent} & ResNet-152 & 46.5 & 73.6 \\
  ACNet~\cite{hu2019acnet} & ResNet-50 & 48.3 & - \\
  Malleable 2.5D~\cite{xing2020malleable} & ResNet-101 & 50.9 & 76.9 \\ 
  SA-Gate~\cite{chen2020bidirectional} & ResNet-101 & 52.4 & 77.9 \\ 
 \hline
  Ours & ResNet-101 & \textbf{53.1} & \textbf{78.1}\\
  \hline
\end{tabular}

\caption{ Comparison with state-of-the art methods on NYU-Depth-v2. Training with InverseForm consistently improves results, and outperforms state-of-the-art methods.}\label{tab:nyud_1}
\end{table}

\begin{table*}[t] 
\centering
 \begin{tabular}{c |c |ccccc |c} 
 \hline
 Network & InverseForm & Seg ($\uparrow$) & Edge ($\uparrow$) & Saliency ($\uparrow$) & Parts ($\uparrow$) & Normals ($\downarrow$) & $\Delta_{m}(\%)$ ($\uparrow$) \\ 
 \hline
 \hline
 
  \multirow{2}{*}{SE-ResNet-26+ASTMT} & & 64.61  & 71.0 & 64.70 & 57.25 & 15.00 &  -0.11\\
  & \checkmark & \textbf{65.13}  & \textbf{71.4} & \textbf{65.29} & \textbf{57.93} & 15.07 & \textbf{0.49}\\ 
 \hline
  \multirow{2}{*}{SE-ResNet-50+ASTMT} & & 68.00  & 72.4 & 66.13 & 61.10 & 14.60 & -0.04\\
  & \checkmark & \textbf{68.83}  & \textbf{72.5} & \textbf{67.50} & \textbf{61.13} & \textbf{14.55} & \textbf{0.95}\\ 
 \hline
  \multirow{2}{*}{SE-ResNet-101+ASTMT} & & 68.51 & 73.5 & 67.72 & 63.41 & \textbf{14.37} & -0.6\\
  & \checkmark & \textbf{70.14} & \textbf{73.7} & \textbf{68.70} & \textbf{64.76} & 14.55 & \textbf{0.39}\\ 
 \hline
  \multirow{2}{*}{ResNet-18+MTI-Net} & & 65.70 & 73.9 & 66.80 & 61.60 & 14.60 & 3.84\\
  &\checkmark & \textbf{65.96} & \textbf{74.2} & \textbf{67.23} & \textbf{61.71} & \textbf{14.52} & \textbf{4.34}\\ 
 \hline 
  \multirow{2}{*}{HRNet-w18+MTI-Net} & & 64.30 & 73.4 & 68.00 & 62.10 & 14.80 & 2.74\\
  &\checkmark & \textbf{65.12} & \textbf{73.6} & \textbf{68.61} & \textbf{62.53} & \textbf{14.67} & \textbf{3.72}\\ 
 \hline

\end{tabular}
    \caption{Training state-of-the-art multi-task learning methods on PASCAL by adding InverseForm loss over boundary detection. HRNet-18 and SE-Resnet backbones are used in a multi-task setting and mIoU scores for segmentation, saliency, human parts and surface normal tasks as well as F-scores for boundary detection are compared with the original results. InverseForm loss consistently improves results barring a few cases.}
\label{tab:pascal} 
\end{table*}

\subsection{Results on PASCAL}
For the PASCAL dataset, we report scores comparing with the state-of-the-art multi-task learning results from ASTMT~\cite{maninis2019attentive} and MTI-Net~\cite{vandenhende2020mti}. As implemented by ASTMT, we use a base architecture of Deeplab-V3+~\cite{deeplab} with a ResNet encoder and squeeze-and-excite (SE)~\cite{squeezeexcite} blocks. We train this network both with and without the InverseForm loss function over the boundary detection task. We also add our loss term over boundary detection task to train the HRNet-w18 backbone from MTI-Net. On this benchmark, we observe consistent improvement in performance by training with our InverseForm loss. These results can be found in Table~\ref{tab:pascal}. We calculate 
the multi-task metric 
relative to the single-task scores reported by the respective papers. The hyperparameters used for our training algorithm are shown in the Appendix.

\begin{figure}[t]
\center
\begin{tabularx}{\textwidth}{c c c}
Input & GSCNN & w/ InverseForm \\
\hspace{-4mm}
\includegraphics[ width=0.33\linewidth, keepaspectratio]{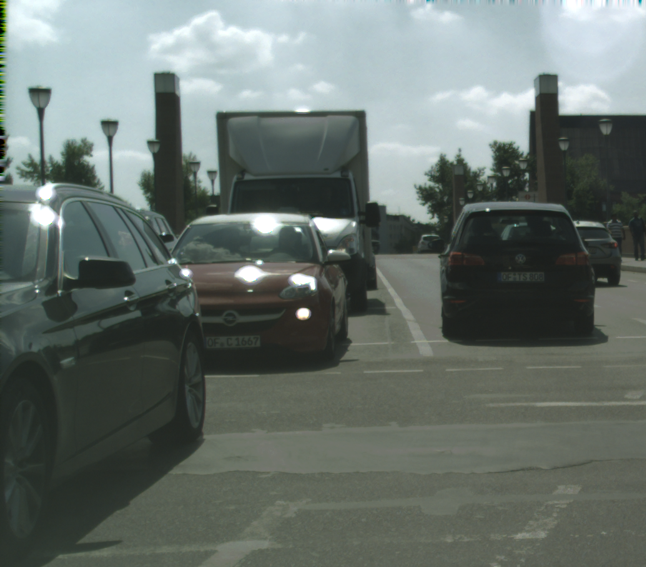} \hspace{-4mm} &
\includegraphics[ width=0.33\linewidth, keepaspectratio]{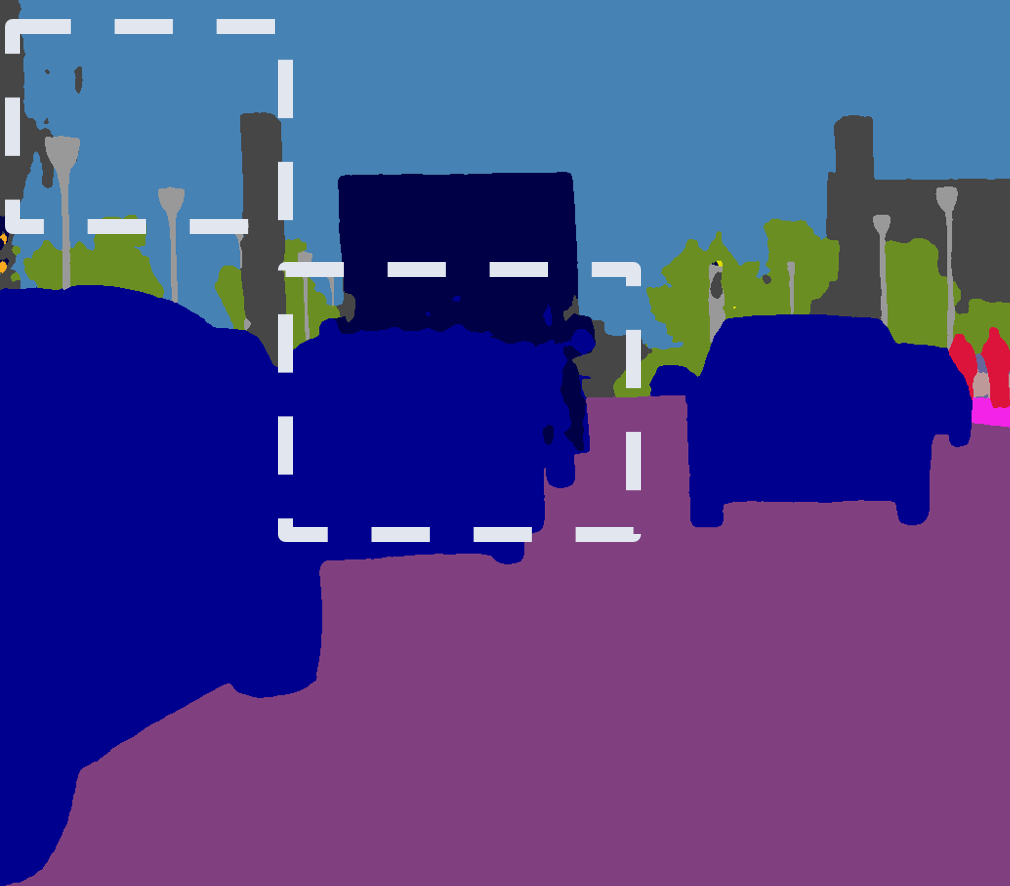} \hspace{-4mm} &
\includegraphics[ width=0.33\linewidth, keepaspectratio]{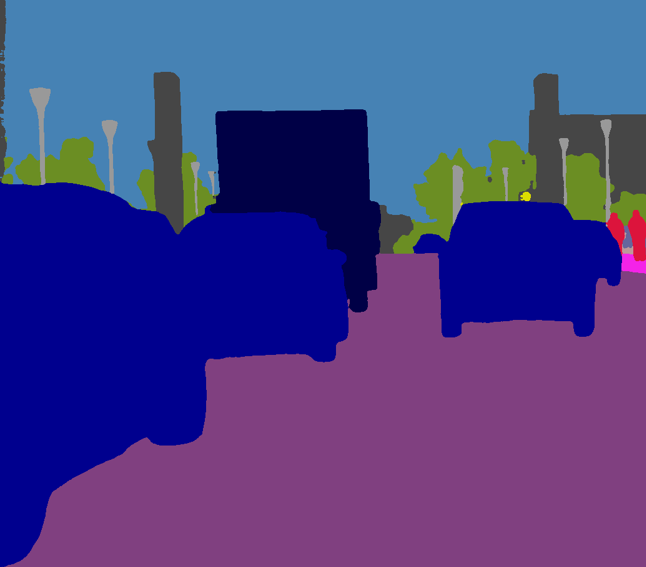} \\

Input & HMS & w/ InverseForm \\

\hspace{-4mm}
\includegraphics[ width=0.33\linewidth, keepaspectratio]{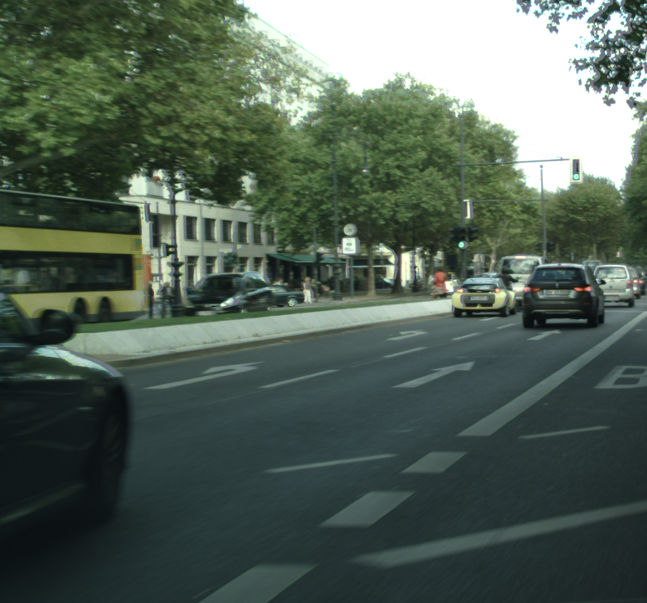} \hspace{-4mm} &
\includegraphics[ width=0.33\linewidth, keepaspectratio]{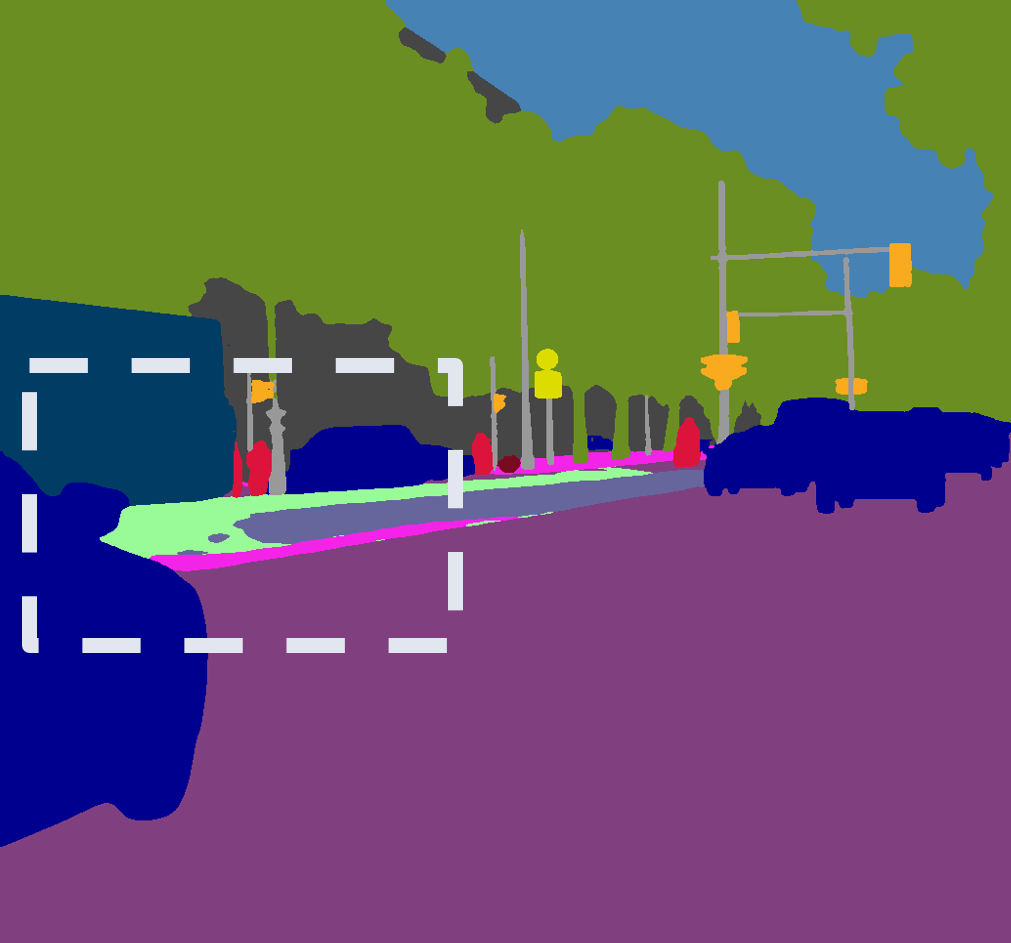} \hspace{-4mm} &
\includegraphics[ width=0.33\linewidth, keepaspectratio]{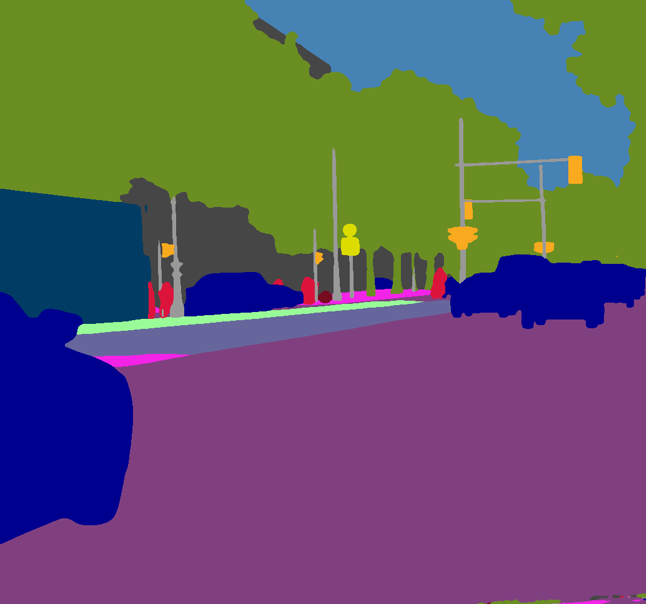} \\

Input & OCR+SegFix &  OCR+InverseForm \\

\hspace{-4mm}
\includegraphics[ width=0.33\linewidth, keepaspectratio]{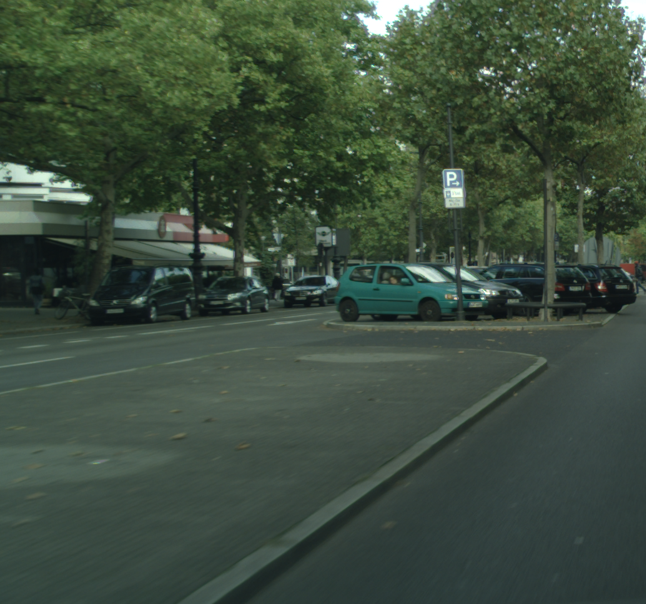} \hspace{-4mm} &
\includegraphics[ width=0.33\linewidth, keepaspectratio]{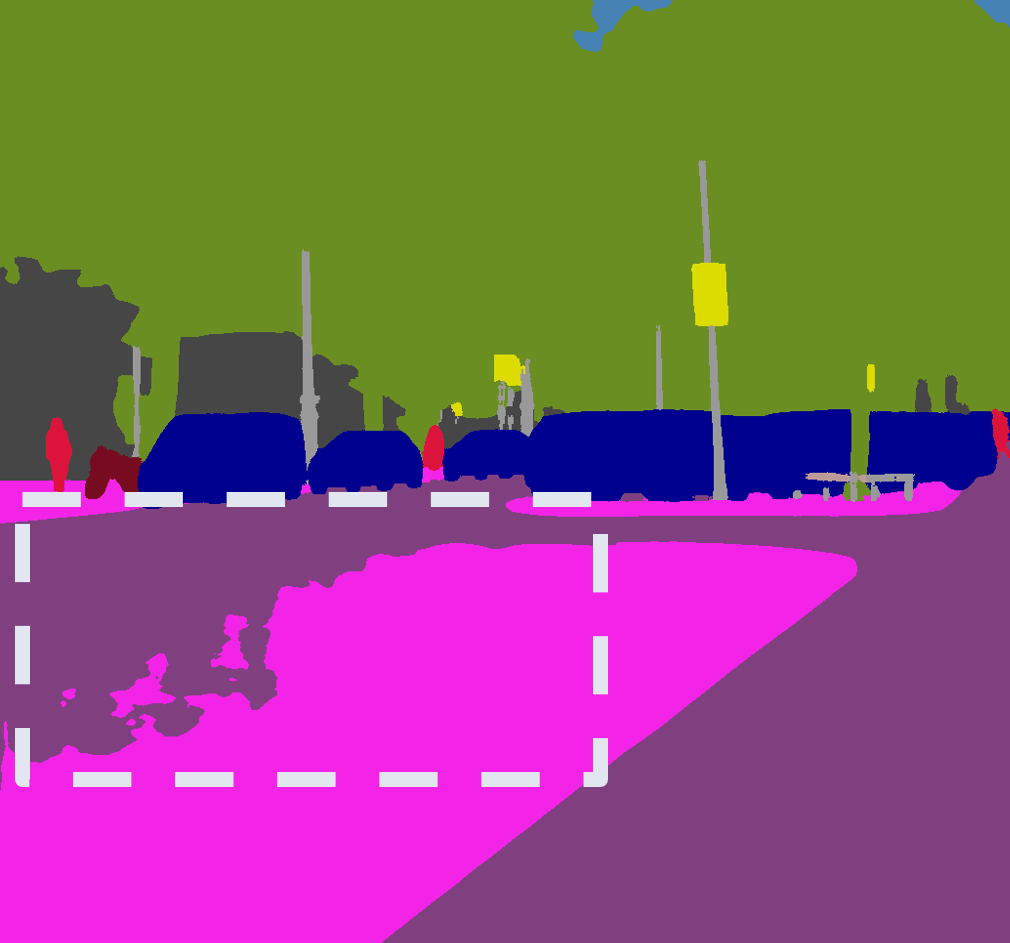} \hspace{-4mm} &
\includegraphics[ width=0.33\linewidth, keepaspectratio]{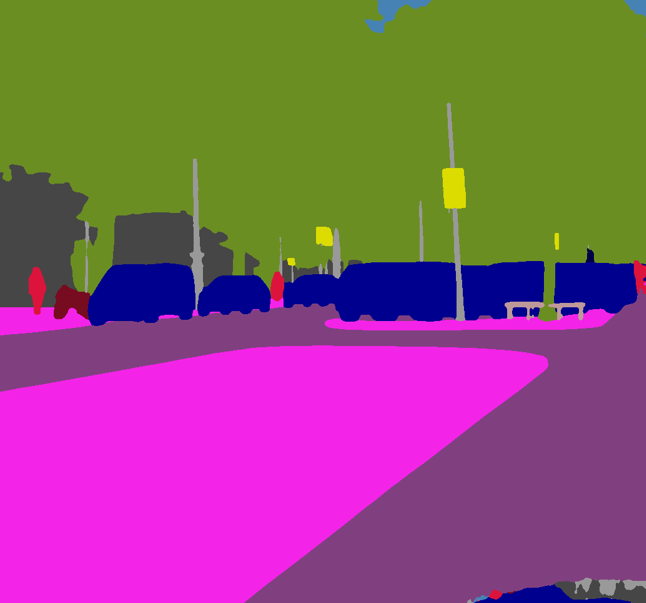} \\

\end{tabularx}
\caption{Cityscapes results showing visual effect of training different baselines with InverseForm loss. The structure of predicted outputs is improved in highlighted regions due to boundary-aware segmentation.} \label{fig:cityscapes results}
\end{figure}

\subsection{Results on Cityscapes}
We report results on the Cityscapes \textit{val} and \textit{test} splits. As a baseline, we use an HRNet-48+OCR~\cite{yuan2020objectcontextual} model. We train the model by adding InverseForm both with and without hierarchical multi-scale attention (HMS)~\cite{tao2020hierarchical}. We also use their auto-labeled coarse annotations. We train with our boundary-aware segmentation scheme from Section~\ref{sec:bd_aware_seg}. We only use the backbone during inference, as illustrated in Figure~\ref{fig:overall_framework}; the network does not require any extra compute compared to the baseline.

We compare our approach to state-of-the-art boundary-aware segmentation frameworks such as Gated-SCNN~\cite{takikawa2019gated} and SegFix~\cite{yuan2020segfix}, quantitatively in Table~\ref{tab:new_cityscapes} and qualitatively in Figure~\ref{fig:cityscapes results}. We outperform SegFix using an HRNet-OCR backbone by 0.3$\%$ mIoU, and using a GSCNN model with WideResNet-38~\cite{zagoruyko2017wide} backbone by 1.1$\%$ mIoU. For GSCNN, we show that our method is complimentary to their work. We train their network by adding the InverseForm loss term to their existing pixel-based boundary loss, outperforming their baseline by 1.6$\%$ mIoU. In Table~\ref{tab:new_cityscapes}, all networks are pretrained on Mapiliary Vistas~\cite{mapiliary}. They use different training sets provided by Cityscapes. These contain coarse and finely-annotated training samples, which are available in the dataset. In Figure~\ref{fig:cityscapes results}, we use the images provided by authors of ~\cite{yuan2020segfix} and ~\cite{takikawa2019gated}, and use the state-of-the-art model provided by ~\cite{tao2020hierarchical} to generate baseline predictions. We then use the same baselines trained with InverseForm from Table~\ref{tab:new_cityscapes} to generate our predictions. As seen in Table ~\ref{tab:new_cityscapes} and Figure~\ref{fig:cityscapes results}, all models trained using InverseForm loss consistently improve compared to their baselines. We are rank 3 on the public Cityscapes leaderboard at the time of submission.

\begin{table}[t]
\centering
        \begin{tabular}{cc |ccc |c}
            \hline
            Method &Backbone &Split &F &C &mIoU \\
            \hline\hline
            Naive-student~\cite{chen2020naivestudent} & WRN41 &Test &\checkmark &\checkmark & 85.2 \\

            GSCNN~\cite{takikawa2019gated} &WRN38 &Test &\checkmark & & 82.8 \\
            HRNet-OCR~\cite{yuan2020segfix} &HRNet48 &Test &\checkmark & \checkmark & 84.2 \\
            OCR+SegFix~\cite{yuan2020segfix} &HRNet48 &Test &\checkmark & \checkmark & 84.5 \\
            OCR \inv{} &HRNet48 &Test &\checkmark & \checkmark & 84.8 \\            
            HMS~\cite{tao2020hierarchical} & HRNet48 &Test &\checkmark &\checkmark & 85.1\\
            HMS \inv{} & HRNet48 &Test &\checkmark &\checkmark & \textbf{85.6}\\\hline
            
            GSCNN~\cite{takikawa2019gated} &WRN38 &Val &\checkmark & & 81.0 \\
            GSCNN+SegFix &WRN38 &Val &\checkmark & & 81.5 \\
            GSCNN \inv{} &WRN38 &Val &\checkmark & & 82.6  \\          
            GSCNN \inv{} &WRN38 &Val &\checkmark & \checkmark & 84.0  \\           
            HMS & HRNet48 &Val &\checkmark &\checkmark &86.7\\
            HMS \inv{} & HRNet48 &Val &\checkmark &\checkmark & \textbf{87.0}\\\hline
        \end{tabular}
    \caption{Our method compared to various state-of-the-art algorithms on Cityscapes. The models reporting on test split are trained using training+validation data. F:Fine annotations, C:Coarse annotations}
    \label{tab:new_cityscapes}
\end{table}

\begin{figure}[t]
\center
\begin{tabularx}{\textwidth}{c c c c}
Input & GT & w/o Ours & w/ Ours \\
\hspace{-4mm}
\includegraphics[ width=0.25\linewidth, keepaspectratio]{nyu_194.jpg} \hspace{-4mm} &
\includegraphics[ width=0.25\linewidth, keepaspectratio]{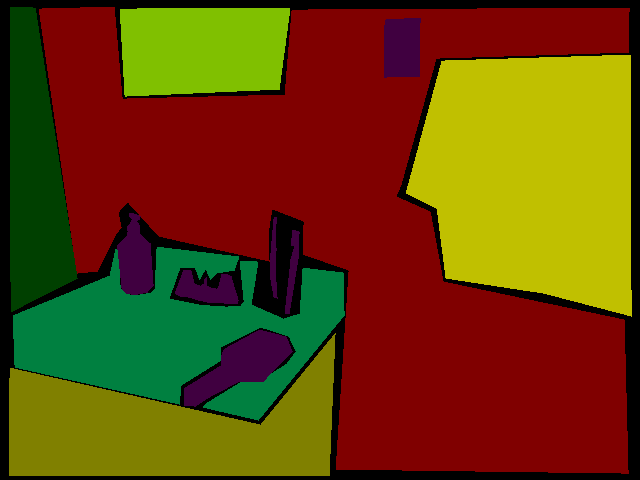} \hspace{-4mm} &
\includegraphics[ width=0.25\linewidth, keepaspectratio]{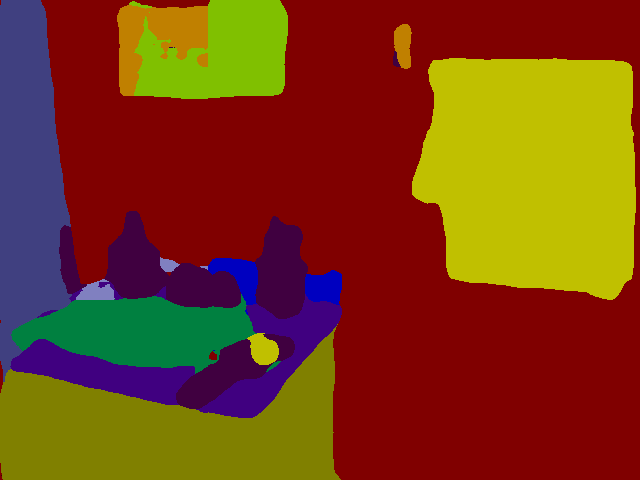} \hspace{-4mm} &
\includegraphics[ width=0.25\linewidth, keepaspectratio]{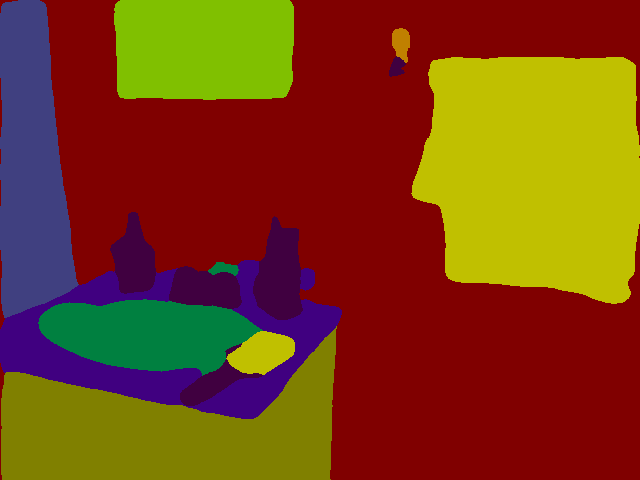} \\
\hspace{-4mm}
\includegraphics[ width=0.25\linewidth, keepaspectratio]{nyu_411.jpg} \hspace{-4mm} &
\includegraphics[ width=0.25\linewidth, keepaspectratio]{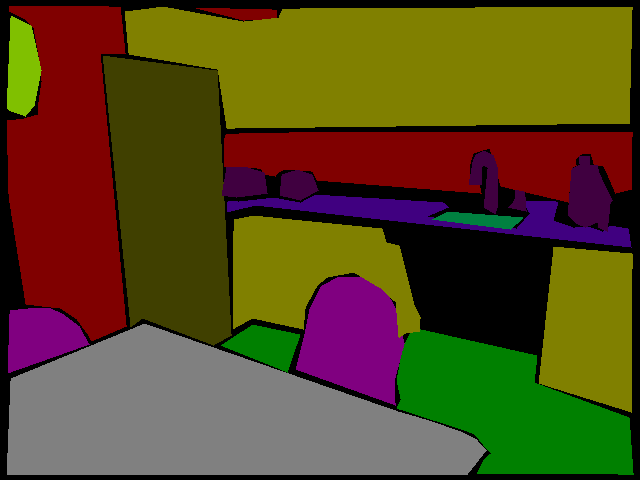} \hspace{-4mm} &
\includegraphics[ width=0.25\linewidth, keepaspectratio]{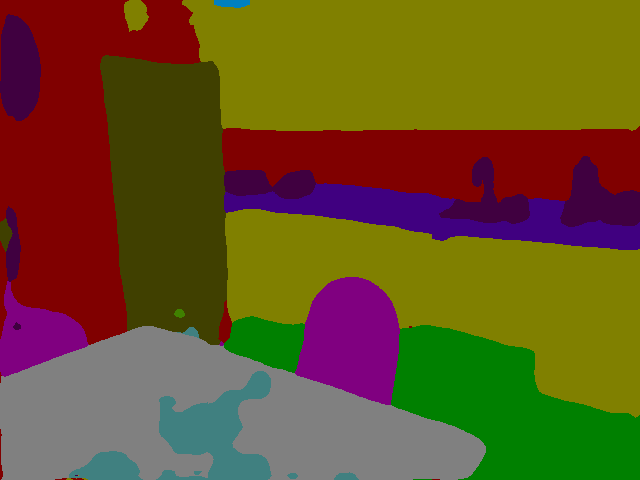} \hspace{-4mm} &
\includegraphics[ width=0.25\linewidth, keepaspectratio]{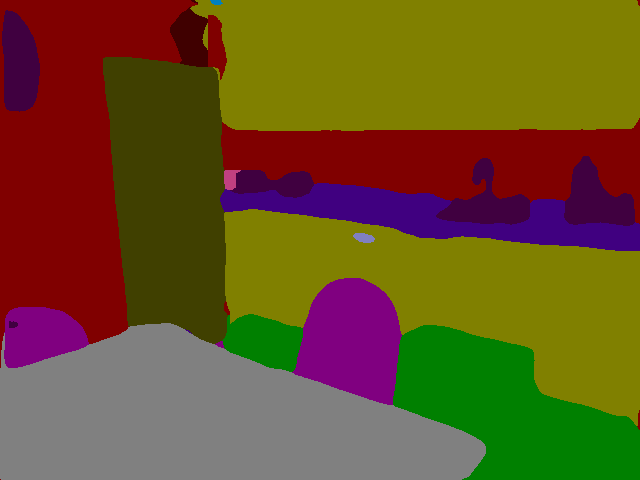} \\
\hspace{-4mm}
\includegraphics[ width=0.25\linewidth, keepaspectratio]{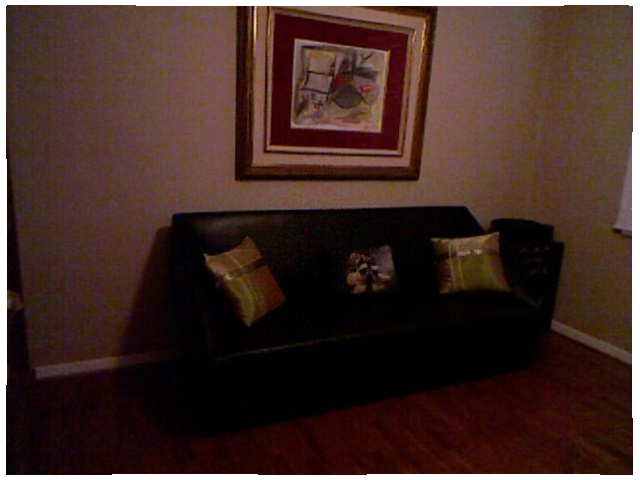} \hspace{-4mm} &
\includegraphics[ width=0.25\linewidth, keepaspectratio]{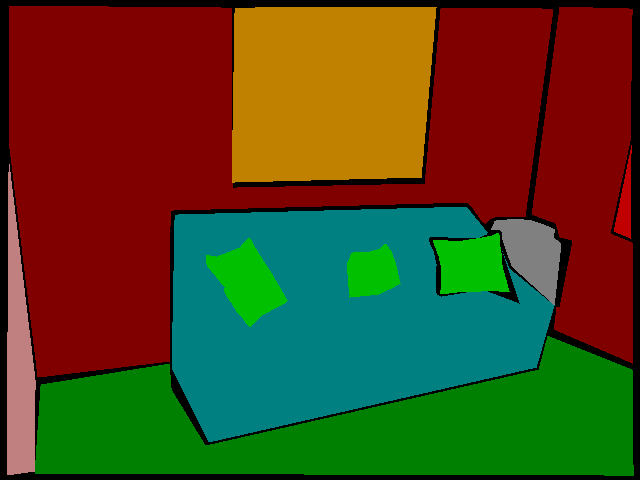} \hspace{-4mm} &
\includegraphics[ width=0.25\linewidth, keepaspectratio]{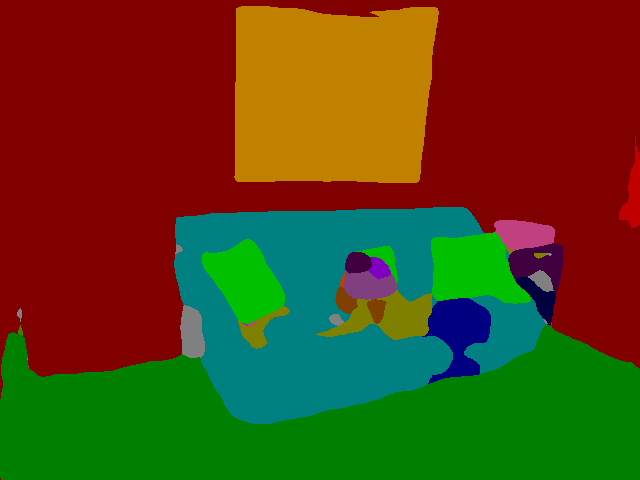} \hspace{-4mm} &
\includegraphics[ width=0.25\linewidth, keepaspectratio]{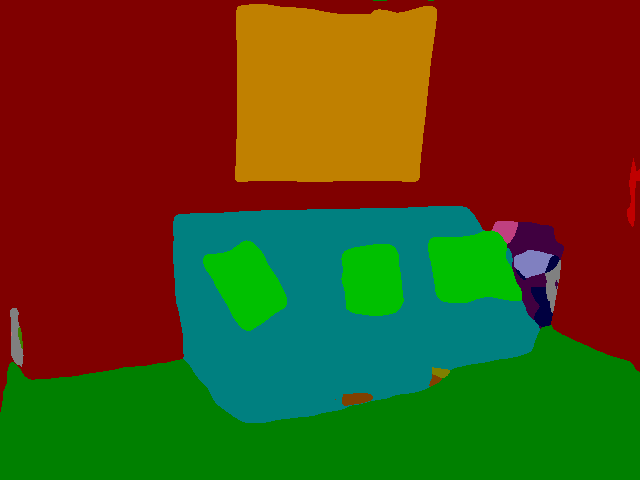} \\

\end{tabularx}
\caption{NYU-Depth-v2 results showing visual effect of training SA-Gates~\cite{chen2020bidirectional} baseline with InverseForm loss. Clear improvement is visible in the structure of predicted outputs due to boundary-aware segmentation.}\label{fig:nyu_result}
\end{figure}

\section{Ablation Studies} \label{sec:ablation}

\noindent\textbf{Searching for the best inverse-transformation:} \label{sec:ablation_architecture_search} We analyze how our current dense inverse-transformation architecture performs compared to the convolutional architecture used in AET~\cite{zhang2019aet}. To reproduce their result, we use their pretrained model and fine-tune with tiled NYU-Depth-v2 boundary images, using their training strategy. We also conduct experiments with both formulations of distance given in Section~\ref{sec:distancemeasures} on NYU-Depth-v2 using vanilla HRNet-w48 and HRNet-w18 as backbones, and the architecture defined in Figure~\ref{fig:overall_framework}. The loss functions used are InverseForm using geodesic distance measure $L_{if, geo}$ and InverseForm using the Euclidean measure $L_{if, euc}$.
Our experiments show a clear improvement in performance for the inverse-transformer architecture v/s the one used by AET. We hypothesize that the reason for better performance is that convolutions are translation-invariant. While the framework used by AET stacks two convolutional embeddings, the local resolution takes a hit compared to global context. 
\begin{table}[h]
\centering
        \begin{tabular}{c| ccc| c}
            \hline
            model &Loss net. &$L_{if, geo}$ &$L_{if, euc}$ &mIoU \\
            \hline\hline
            \multirow{5}{*}{HRNet-w48} & AET &\checkmark & & 47.19 \\
                                    & Ours &\checkmark & & 47.28 \\
                                    & AET & &\checkmark & 47.03 \\
                                    & Ours & &\checkmark & \textbf{47.42} \\\hline
            \multirow{5}{*}{HRNet-w18} & AET &\checkmark & & 33.82 \\
                                    & Ours &\checkmark & & \textbf{34.84} \\
                                    & AET & &\checkmark & 33.97 \\
                                    & Ours & &\checkmark & 34.79 \\\hline        
    \end{tabular}
    \caption{Searching for the optimal InverseForm loss}
    \label{tab:architecture_search}
\end{table}

\noindent\textbf{Distance function:} Our results for using the geodesic distance v/s Euclidean distance however, do not show a clear winner. This is partly due to the reason that the formulation in Section~\ref{sec:distancemeasures} for geodesic distance can lead to exploding gradients easily. This severely limits the search-space for hyperparameters. Euclidean distance on the other hand might not model perspective homography best, but using this gives us a wider search-space and hence a more consistent improvement. We recognize the need to explore the benefits of using geodesic distance, and hence continue our research on the optimal InverseForm network architecture.

\section{Conclusion} \label{sec:conclusion}
We propose a distance-based boundary-aware segmentation method that consistently improves any semantic segmentation backbone it is plugged in. During training, the boundary detection and segmentation are jointly optimized in a multi-task learning framework. At inference time, it does not require any additional computational load. The distance-based measure outperforms cross-entropy based measures while efficiently capturing the boundary transformation. We show empirically that our scheme achieves superior segmentation accuracy and better structured segmentation outputs. We continue to look for the optimal architecture and distance measure for this method, as we show some room for improvement in our ablation experiments.

{\small
\bibliographystyle{ieee_fullname}
\bibliography{egbib}
}

\end{document}